\documentclass[runningheads,a4paper]{llncs}

\usepackage{amssymb}
\usepackage{graphicx}
\usepackage{graphicx}
\usepackage{tabularx}
\usepackage{soul}

\usepackage{epstopdf}
\usepackage[latin1]{inputenc}

\usepackage{hyperref}
\usepackage{xstring}

\usepackage{times}
\usepackage{latexsym}
\usepackage{url}
\usepackage{amssymb}
\usepackage{amsmath}
\usepackage{latexsym}
\usepackage{graphicx}
\usepackage{algpseudocode}
\usepackage[ruled]{algorithm2e}
\usepackage{enumitem}
\usepackage{multirow}
\usepackage{pifont}
\usepackage{todonotes}
\usepackage{hyphenat}

\urldef{\mailsa}\path|{alfred.hofmann, ursula.barth, ingrid.haas, frank.holzwarth,|
\urldef{\mailsb}\path|anna.kramer, leonie.kunz, christine.reiss, nicole.sator,|
\urldef{\mailsc}\path|erika.siebert-cole, peter.strasser, lncs}@springer.com|

\begin{document}

\mainmatter  

\title{
Combining Graph-based 
Dependency Features with 
Convolutional Neural Network for Answer Triggering
}

\author{Deepak Gupta$^{\ast}$, Sarah Kohail$^{\dagger}$,  
Pushpak Bhattacharyya$^{\ast}$}

\institute{
$^{\ast}$Indian Institute of Technology Patna, India\\ $^{\dagger}$Universit{\"a}t Hamburg, Germany \\
         \{deepak.pcs16, pb\}@iitp.ac.in\\
         \{kohail\}@informatik.uni-hamburg.de\\}

%
%


%
%

\maketitle

\begin{abstract}
Answer triggering is the task of selecting the best suited answer for a given question from a set of candidate answers, if exists.
In this paper, we present a hybrid deep learning model for answer triggering, which combines several dependency graph based alignment features, namely graph edit distance, graph based similarity and dependency graph coverage, with dense vector embeddings from a Convolutional Neural Network (CNN).  
Our experiments on the WikiQA dataset show that such a combination can more accurately trigger a candidate answer compared to the previous state-of-the-art models. Comparative study on \textsc{WikiQA} data set shows $5.86\%$ absolute F-score improvement at the question level.
\end{abstract}

\section{Introduction} 

Answer triggering is a relatively new problem for open-domain question answering (QA). In addition to extracting correct answers from a set of pre-selected candidate pool (i.e answer selection), answer triggering detects whether a correct answer exists in the first place \cite{yang-yih-meek:2015:EMNLP,severyn2013automatic,heilman2010tree}. \\
\indent
To evaluate the performance of answer sentence selection, \textsc{WikiQA} dataset has been widely used.
It consists of questions collected from the user logs of the Bing search engine. The dataset is constructed using a more natural process and it also includes questions for which there exists no correct answer. Lexical similarity between a question and answer pair in the \textsc{WikiQA} dataset is also lower 
as compared to other answer sentence selection datasets like the dataset provided by TREC QA\footnote{\url{https://trec.nist.gov}} and \textsc{QASent}.
In some cases, there is no lexical overlap at all, as shown in the following example. Given a question $\textbf{Q}$ and correct answer $\textbf{A}$. $\textbf{Q}$ and $\textbf{A}$ does not follow any lexical similarity.
\\
\noindent
\textbf{Q:} \textit{what can sql 2005 do ?} \\
\textbf{A:} \textit{As a database, it is a software product whose primary function is to store and retrieve data as requested by other software applications.
 }\\
This makes the answer triggering task more challenging than usual answer sentence selection. \\ Applying deep-neural-network-based models has shown a significant progress in the absence of lexical overlap between question and answer \cite{yao2013answer,DBLP:journals/corr/WanLGXPC15,DBLP:journals/corr/FengXGWZ15}. However, such models still ignore the importance of grammatical and structural relations in the context of this task. 

In this paper, we propose an effective model for answer triggering, 
which \textbf{(i)} detects whether there exists at least one correct answer in the set of candidate answers for a given target question, in the absence of explicit lexical overlap, and \textbf{(ii)} finds the most appropriate answer from a set of candidate answers, if there exists any such. Our contribution handles both -- fuzzy lexical matching via Convolutional Neural Network (CNN) and grammatical structure matching via encoding dependency graphs overlap. The CNN can capture the semantic similarity between question and answer whereas dependency graph-based features capture structural overlap resp. divergence where lexical similarity is high.
We perform experiments on \textsc{WikiQA} dataset \cite{yang-yih-meek:2015:EMNLP}  and show that introducing graph-based features into the CNN performs superior as compared to CNN alone, significantly outperform previous state-of-the-art methods.
\section{Related Work}
The complexity of question answering research has been increased in recent years. Due to the wider success of deep learning based model in other NLP area such as named entity recognition (NER) \cite{dos2015boosting,kumar2016recurrent,GUPTA18.486,2016deep}, sentiment analysis \cite{socher2013recursive,gupta-EtAl:2016:W16-63,lrec2018} and parsing \cite{socher2013parsing,turian2010word}, several deep learning based model \cite{dong2015question,xiong2016dynamic,ren2015exploring} has been used to solve the Q/A problem.
learn to match questions with answers by two model bag-of-words and biagram using convolutional neural network with a single convolution layer, average pooling and logistic regression.
\cite{iyyer-EtAl:2014:EMNLP2014} present \textit{qanta}, a dependency-tree recursive
neural network for factoid question answering which effectively learns word and phrase-level representations. 
Convolutional neural network based deep learning model is very popular in Q/A, its success has been reported by \cite{yu2014deep,TACL831,yih-he-meek:2014:P14-2,GUPTA18.826,kalchbrenner-grefenstette-blunsom:2014:P14-1,yin2016attention}. Recently deep neural variational inference \cite{miao2016neural} present for answer sentence selection.
\cite{yang-yih-meek:2015:EMNLP} and \cite{selQAPaper} proposed a CNN based model for answer triggering. However our CNN inspired model differ from them to calculate the semantic similarity between question and answer by means of  dependency graph similarity, matching and coverage.
\section{Hybrid Model for Answer Triggering}\label{cnn}
Our method uses Convolutional
Neural Network (CNN) to extract deep generic features from question-answer pairs. 
Our CNN maps each input sentence into a vector space, preserving syntactic and semantic aspects, which enables it to generate an effective and diverse set of features \cite{kim2014convolutional,collobert2011natural,kalchbrenner2014convolutional}. 
\subsection{Convolutional Neural Network (CNN) Model}
A simple CNN model takes a sentence as an input and performs convolution followed by pooling and classify the sentence into one of the predefined classes by a soft-max classifier. A Joint-CNN is an advancement where question and candidate answer are both input to the model. The convolution and pooling operations for questions and answers are performed separately. Thereafter, the outputs of fully connected layers (for question and answer) are concatenated to form a single input to the soft-max classification layer. The Joint-CNN model provides probabilistic score as an output. It is inspired by the Yoon Kim \cite{kim2014convolutional} CNN architecture for text classification. We describe model components in the following.
\paragraph{Question/Answer Representation Matrix.}
Given a question $Q$ and candidate answer $A$, having $n_Q$ and $n_A$ number of token respectively, each token $t_i \in Q$ and $t_j \in A$ is represented by $k$ dimension 
distributed representation $\textbf{$x$} \in \mathbb{R}^{k} $ and $\textbf{$y$} \in \mathbb{R}^{k}$ respectively.
%
The question and answer representation matrices 
can be generated by concatenating column level,  $1 \times k$ dimensional word vector to form a $n_Q \times k$ and $n_A \times k$ dimensional matrix. 
We set a maximum number of tokens\footnote{We set maximum $20$ and $40$ tokens for question and answer sentence, respectively} to create question and answer matrix.
\paragraph{Convolution and Pooling}
In order to extract common patterns 
throughout the training set, we use the convolution operation using different feature detector (filter) length \cite{kim2014convolutional} on the question/answer representation matrix. We also apply the max pooling operation over the feature map similar to \cite{collobert2011natural} on convolution output of both question and answers. In our experiment we used two layer of convolution followed by pooling layer.
\paragraph{Fully Connected Layer}
Finally, outputs of pooling layers $p_Q$ and $p_R$ are concatenated, and this resulting pooling layer $p=p_Q \otimes p_A$ is subjected to a fully connected softmax layer. 
It computes the probability score over two label-pair \textit{viz} trigger, non-trigger as:
\begin{equation} 
\begin{split}
P(c=l|Q,A,p,a) & =softmax_{l}(p^{T}\textbf{w}+a)\\
 & = \frac{e^{p^{T}\textbf{$w_l$}+a_l}}{\sum_{k=1}^{K}e^{p^{T}\textbf{$w_k$}+a_k}}
\end{split}
\end{equation}
where $a_k$ and $w_k$ represent the bias and weight vector, respectively, of the $k^{th}$ label.
\begin{figure*}[h]
\centering
\includegraphics[height=10cm,width=\linewidth]{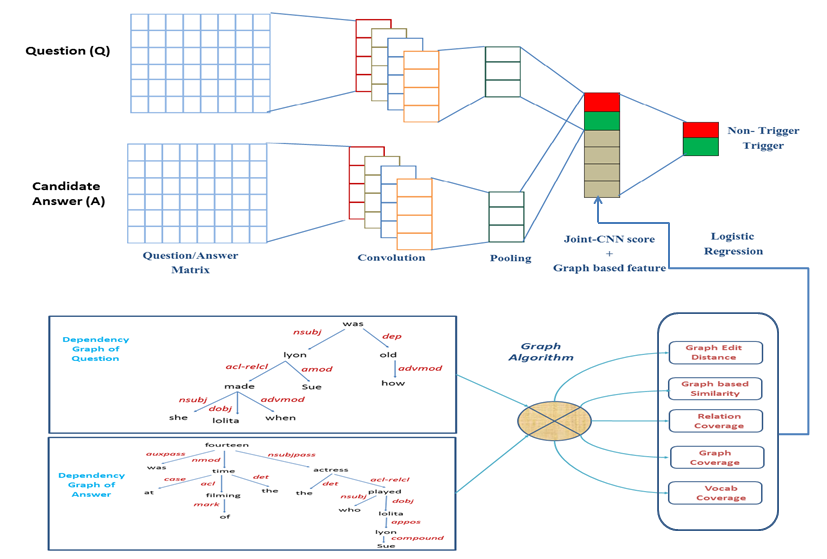}
 \caption{Proposed model architecture for answer triggering. The architecture combines mainly two component Joint-CNN and dependency graph based alignment features. Both component works independently by taking a question and answer as input. Logistic regression is used to predicted the final label, either \textit{`Trigger'} or \textit{`Non-Trigger'}.  }
  \label{fig:model}
\end{figure*}

\subsection{Dependency Graph-based Features}
\begin{figure*}[!tbh]
\small
\centering
\includegraphics[]{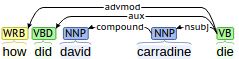}
\quad
\includegraphics[width=\linewidth]{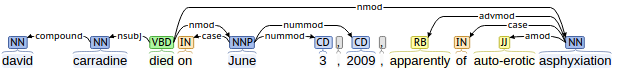}
 \caption{Dependency graph of a question \textbf{Q:}\textit{``how did david carradine die''} and their correct answer \textbf{A:}\textit{``david carradine died on June 3 , 2009 , apparently of auto-erotic asphyxiation''}. The word \textit{`david'} and \textit{`carradine'} have the same dependency relation \textit{`compound'}. The dependency link between word \textit{`die'} and \textit{`how'} in question and \textit{`die'} and \textit{`asphyxiation'} in answer provide the similarity and coverage between question and answer.}
 \label{dep-graph}
\end{figure*}

Both question and answer are converted into a graph using the dependency relations obtained from the Stanford 
dependency parser\footnote{http://nlp.stanford.edu:8080/parser/}, following \cite{kohail2015unsupervised}. 
Dependency graphs of question and answer share common subgraphs of 
dependency links between words (c.f. Fig-\ref{dep-graph} for an example).
Based on these graphs, we extract three sets of features: graph edit distance, similarity features and coverage features.
\subsubsection{Graph Edit Distance}
Graph edit distance defines the cost of the least expensive sequence of edit operations that are needed to transform one graph, in our case dependency parsing tree, into another. It calculate the minimum cost required to transform the question graph to an answer graph. Table-\ref{gedit} shows the effectiveness of this feature to determine the correct answer from the pool of candidate answers. We calculate the node and edge difference between the dependency graph of question and answer. In node difference, if the two nodes from question and answer have same word (lemma) then node difference will be `0'. Given the different word, the node  difference is calculated by a parts of speech (POS) substitute weight . The difference between two POS-tags is measure by substitute weight. For instance, replacing a noun with a verb should be more costly than replacing a verb with a verb. Similar for node difference the edge (dependency relation) difference between question and answer is  calculated. A two-dimensional cost matrix can be created, by considering the graph edit distance between question and answer, which represents the cost of each possible node edit operation. Finally the optimal cost are obtained by assignment algorithm \cite{munkres1957algorithms}.
\begin{table*}
\centering
\caption{Graph-edit distance between a question and candidate answer pair. The answer which is in \textbf{bold} is the correct answer for question. }
\resizebox{\linewidth}{!}{
\begin{tabular}{l|l|r} \hline
Question & Candidate Answer & Graph-Edit Distance \\ \hline
\multirow {3}{*}{how old was sue lyon when she made lolita?} 
& \begin{tabular}[l]{@{}l@{}}Lolita is a 1962 comedy-drama film by Stanley\\Kubrick based on the classic novel of the same\\title by Vladimir Nabokov.\end{tabular} & 0.98 \\ \cline{2-3}
& \textbf{\begin{tabular}[l]{@{}l@{}}The actress who played Lolita, Sue Lyon , \\ was fourteen at the time of filming\end{tabular}} & \textbf{0.59} \\ \cline{2-3}
& \begin{tabular}[l]{@{}l@{}}Kubrick later commented that, had he realized \\ how severe the censorship limitations were going\\to be, he probably never would have made the film\end{tabular} & 0.71 \\ \hline
\end{tabular}%
}

\label{gedit}
\end{table*}
\subsubsection{Dependency Graph based Similarity}
For each sentence\footnote{Sentence is either question or answer} $S$, we define the dependency graph $G_S$ = $\{V_S,E_S\}$, where $V_S$ = $ \{t_1,t_2,\ldots,t_{n_S}\}$ represent the tokens in a sentence, and $E_S$ is a set of edges. Each edge $e_{ij}$ represents a directed dependency relation between $t_i$ and $t_j$.
We calculate TF-IDF \cite{salton1988term} three levels and weight our dependency graph using the following conditions:\\
 \textbf{Word TF-IDF}: Consider only those words that satisfy a criteria $\alpha_1$. 
TF-IDF ($S$, $t_i$) $> \alpha_1$\\ 
 \textbf{Pair TF-IDF}: Word pairs are filtered based on the criteria $\alpha_2$.
TF-IDF ($S$, $t_i$, $t_j$) $> \alpha_2$\\
 \textbf{Triplet TF-IDF}: Consider only those triplet (word, pair and relation), which satisfies a condition $\alpha_3$.
TF-IDF ($S$, $t_i$, $t_j$, $e_{ij}$) $> \alpha_3$\\
Similarities are then measured on three levels by representing each sentence as a vector of words, pairs and triples, where each entry in one vector is weighted with the TF-DF measure. The IDF is computed using the NYT part of the Gigaword corpus
\cite{graff2003english}.
\subsubsection{Dependency Graph-based Coverage}
To overcome the bias of higher similarity values between longer sentences \cite{albalate2013semi}, we use the coverage score between the dependency graphs of question and answer. Let $G_Q$ = $\{V_Q,E_Q\}$ and $G_A$ = $\{V_A,E_A\}$ be the dependency graphs of a pair of question and candidate answer. Intuitively, coverage models the fraction of the question that the answer addresses.
\paragraph{Relation Coverage.}  We compute the number of one-to-one edge correspondence between dependency graph of question $Q$ and answer $A$, divided by the total number of edges in the dependency graph of question $Q$.
\paragraph{Graph Coverage.} The idea is to find a subgraph $G_{Sub}$ in the candidate answer dependency graph $G_A$ that is similar to a given query text dependency graph $G_Q$. We use the dependency sub-graph approximate alignment algorithm by \cite{kohail2017cicling}. The pseudo-code of algorithm is listed in Algorithm \ref{algo}. 
The algorithm obtains the common set of nodes between $G_Q$ and $G_A$ and finds the shortest path between every pair of nodes belongs to the intersection set in the candidate answer dependency graph using Dijkstra's algorithm \cite{dijkstra1959note}. Each edge is assigned to a weight of 1 and edges directions are ignored during the
process of the algorithm. A threshold parameter $t$ is defined, which allows for node gaps and mismatches in the case where some nodes in the answer text cannot be mapped to any nodes in the question graph. If the shortest path size (i.e number of edges between
a pair of nodes) is less than or equal t, the path will be added to the sub-graph Gs. There are two coverage features computed on the sub-graph. 
\begin{itemize}
\item Ratio of relation overlap in sub-graph with respect to answer sentence dependency graph.
\item Ratio of relation overlap in sub-graph with respect to question sentence dependency graph.
\end{itemize}
 \paragraph{Vocabulary Coverage.} We compute the number of one-to-one node correspondence between dependency graph of question $Q$ and answer $A$, divided by the total 
 number of nodes in the dependency graph of question $Q$.
 \begin{algorithm}[!t]\label{algo}
\KwIn{Dependency graph $G_Q$ and $G_A$ and threshold $m$}
\KwOut{Dependency sub-graph $G_{Sub}$}
\Begin{
$V_{common} \leftarrow \{V_{Q} \cap V_{A} \}$ ; \\ 
\For{$i = 1$ to $|V_{common}|-1$}
{
  \For{$j = i+1$ to $|V_{common}|$}
{
   $NodePath= FindPath(G_A, t_i,t_j)$ \;
   \If{$NodePath \not=\phi \wedge size(NodePath, t_i, t_j) \leq m$}{
 $G_{Sub} \leftarrow G_{Sub} \cup NodePath$ \;
   }
} 
}
 }
return $G_{Sub}$
\caption{Pseudo-code of Dependency sub-graph approximate alignment}
\end{algorithm}
 
 \begin{algorithm}[h]\label{algo_supporting}
{
\texttt{procedure:\textbf{FindPath(Graph $G_A$, Vertex $s$, vertex $d$)}}\\
\Begin{
\texttt{computePath(Graph $G_A$, Vertex $s$)}\\
\texttt{FindPathTo(Graph $G_A$, Vertex $d$)}\\
}
}
\caption{}
\end{algorithm}
\begin{algorithm}[h]\label{algo-2}
{
\texttt{procedure:\textbf{computePath(Graph $G_A$, Vertex $s$)}}\\

\Begin{
\For{each vertex v $\in V_A$ }
{
$v.minDistance=\infty$\\
$v.parent=NULL$
}
$s.minDistance = 0 $\\ 
Initialize a priority queue \texttt{vertexQueue}\\
\texttt{vertexQueue}.$add(s)$ \\ 
\While{\texttt{vertexQueue} is not empty}
{
$u$=\texttt{vertexQueue}.$removeHead()$;
\For{each edge e=(u,v)}
{
$temp=u.minDistance+1$ \;
 \If{$temp<v.minDistance$}{
 \texttt{vertexQueue}.$remove(v)$\\
 $v.minDistance=temp$\\
 $v.parent=u$\\
 \texttt{vertexQueue}.$add(v)$
  }
}
}
 }
 }

\caption{Pseudo-code of calculating shortest path from source $s$ in candidate  dependency graph $G_A$}
\end{algorithm}

\begin{algorithm}[h]\label{algo-3}
{
\texttt{procedure:\textbf{FindPathTo(Graph $G_A$,Vertex $d$)}}\\
\Begin{
\begin{small}
Initialize a path list from source to destination $d$\\
\end{small}
$\texttt{path}=\{\}$\\ 
$vertex=d$\\

\While{\texttt{vertex} is not empty}
{
$\texttt{path}.add(vertex)$\\
$vertex=vertex.parent$\\

 }
 $\texttt{path}.reverse()$
 
 }
 return \texttt{path}
}
\caption{Pseudo-code of finding the  shortest path from source to destination }
\end{algorithm}
\section{Datasets, Experiments and Analysis}
\subsection{Datasets and Experimental Setup}
We use the \textsc{WikiQA} data set for our experiments. Statistics of question and answer pairs of \textsc{WikiQA} data set are given in Table \ref{data-stat}. For training the Joint-CNN model as discussed in Section \ref{cnn}, we employ stochastic gradient descent (SGD) 
over mini-batch, and back-propagation \cite{hecht1989theory} to compute the gradients. For word embeddings we use the pre-trained Google word embedding model\footnote{https://code.google.com/archive/p/word2vec/}.
The Ada-delta \cite{zeiler2012adadelta} update rule is used to tune the learning rate. The optimal hyper-parameters\footnote{feature maps size=$100$, drop-out rate=$0.5$, maximum epochs=$50$, learning rate=$0.2$, filter window size=$3,4$, $\alpha_1= 7$, $\alpha_2=5$, $\alpha_3= 2$, $m=2$} are determined on the development data. In our final model, 
we embed probabilistic scores obtained from CNN along with graph based linguistic features to train a logistic regression classifier. The proposed model architecture depicted in fig \ref{fig:model}.
\begin{table}
\centering
\caption{Statistics of \textsc{WikiQA} data}
\begin{tabular}{c|l|l|l}
\hline
 & Train & Dev & Test \\ \hline
No. of Questions & 2,118 & 296 & 633 \\ \hline
No. of Answers & 1,040 & 140 & 293 \\ \hline
No. of Question w/o Answer & 1,245 & 170 & 390 \\ \hline
\end{tabular}%
\label{data-stat}
\end{table}
\subsection{Baselines}
We evaluate the model using information retrieval (IR) and semantic composition based similarity. The following baselines are the   used to evaluate the answer sentence selection and answer triggering task.

\indent
\begin{itemize}
\item \textbf{Baseline-1}: The first baseline is constructed based on the similarity measure using Okapi BM25 algorithm \cite{robertson1995okapi}. Each candidate answer is treated as a single document. We calculate the BM25 score between question $Q$ and a candidate answer $A$. The score of a candidate answer $A$ for a given question $Q$ consisting of the words $q_1, ..., q_n$ is computed as:
\begin{equation} \label{eq:bm25}
\text{Score}(Q,A) = \sum_{i=1}^{n} \text{IDF}(q_i) \cdot \frac{f(q_i, A) \cdot (k_1 + 1)}{f(q_i, A) + k_1 \cdot (1 - b + b \cdot \frac{|A|}{\text{avgdl}})} \end{equation}
where $f(q_i, A)$ is $q_i$'s term frequency in the candidate answer $A$, $|A|$ is the length of the candidate answer (in words), and $avgdl$ is the average candidate answer length in the answer pool. $k_1$ and $b$ are the free parameters.
\begin{equation} \label{eq:idf}
\text{IDF}(q_i) = log\frac{N-n(q_i)+0.5}{n(q_i)+0.5}
\end{equation}
where $N$ is the total number of candidate answers in the answer pool, $n(q_i)$ is the number of candidate answers containing $q_{i}$.\\
An optimal threshold value is estimated from the development data. The candidate answer which is having the score above a certain threshold value is set to `1'(triggered) and the rest are set to `0'(non-triggered).
\item \textbf{Baseline-2}: Our second baseline is based on the n-gram coverage between question and answer. We compute the n-gram coverage upto 3-gram. Finally, the n-gram score between a question and an answer is calculated based on the following formula. 
\begin{equation}
NGCoverage(Q,A,n)=\frac{\sum_{ng_{n}\in A}Count_{common}(ng_n)}{\sum_{ng_{n}\in Q}Count_{ques}(ng_n)}
\end{equation}
\begin{equation}
NGScore(Q,A)=\sum_{i=1}^n \frac{NGCoverage(Q,A,i)}{\sum_{i=1}^{n}i}
\end{equation}
We set a threshold value similar to the first baseline. The candidate answer which is having the score above a threshold value is set to `1'(triggered) and the rest are set to `0'(non-triggered).
\item \textbf{Baseline-3}: We perform experiments using two sets of pre-trained deep learning (DL) based word embeddings, 
Google's word2vec embeddings of dimension $300$\footnote{https://code.google.com/archive/p/word2vec/} and GloVe word embeddings, of dimension $100$\footnote{https://nlp.stanford.edu/projects/glove/}. The question/answer vector is computed as follows,
\vspace{-1em}
\begin{equation} 
\texttt{VEC} (S) = \frac{\sum \nolimits_{t_{i} \in S} \texttt{VEC} (t_{i})}{\textit{number of look-ups}}
\end{equation}
where $S$ is question/answer in interest, \textit{number of look-ups} represents the number of words in the question for which word embeddings are available. The cosine similarity between question vector and candidate answer vector are computed.
An optimal threshold value of cosine similarity is estimated from the development data. 
The candidate answer having cosine similarity above the threshold score (0.70) is set to `1' (triggered), and all others are set to `0'(non-triggered).
\\
\end{itemize}
\subsection{Result and Analysis}
Mean Average Precision (MAP) and Mean Reciprocal Rank (MRR) are used to evaluate the performance for answer sentence selection. But both are not suitable for
evaluating the task of answer triggering because these evaluate the relative ranks of correct answers in the candidates of a question. We use the standard precision, recall and F-score to evaluate the answer triggering problem following \cite{yang-yih-meek:2015:EMNLP}. While evaluating, we consider all the  candidate answers that yield the highest model score. If the score is above a predefined threshold then the candidate answer is labeled as a correct answer to the question. The optimal threshold value (0.14) is determined based on the development data. We define three baseline BM-25, N-Gram coverage and 
semantic similarity based model to compare against our proposed model. 

We conduct experiments with different feature map sizes for Joint-CNN. 
We also analyze the impact of different graph based features. Detailed comparisons and the impact of these features are reported in Table \ref{result}.
We observe the impact of dependency graph feature in determining the suitable answers from a collection of answer pool. The feature ablation study reveals the importance of each dependency graph feature on validation and test set. However, the similar impact of each feature could not observed in test data set. The final model comprised of the best Joint-CNN model (100-FMap) with graph edit distance, graph similarity and graph coverage.
We observe that Joint-CNN with graph based feature achieves an improvement of $6.17$, $4.21$ and $3.41$ points over the Joint-CNN model (without graph feature) with respect to F-score, MAP and MRR. The best combination is obtained with a CNN that maximizes recall, the graph-based features substantially improve precision for the maximal F-score, MRR and MAP. \cite{yang-yih-meek:2015:EMNLP} 
and \cite{selQAPaper} used the same \textsc{WikiQA} dataset to evaluate their system performance on answer triggering task. 
The \cite{yang-yih-meek:2015:EMNLP} model is based on the augmentation of question class and sentence length feature to CNN. A subtree matching algorithm along with CNN architecture is used in \cite{selQAPaper} to evaluate answer triggering. Our proposed model is different from these state-of-the-art models in terms of investigation of richer linguistic feature (coverage, similarity) and graph based similarity in conjunction with CNN model. 
The values obtained through t-test show that performance improvements in our proposed model over these two state-of-the-art systems are statistically significant ($p<0.05$). In Table \ref{analysis} we provide analysis with proper examples for our two proposed models, \textit{viz.} Joint-CNN and Hybrid. 
\begin{table}[h]
\centering
\caption{Neural network hyper-parameters}
\begin{tabular}{c|c|c}
\hline
\textbf{Parameter} & \textbf{Description}&\textbf{Value}\\ \hline
\textit{$d^x$} & Word embedding dimension & 300  \\ 
\textit{n} & Maximum length of comments & 50  \\
\textit{m} & Filter window sizes & 3,4  \\ 
\textit{\textbf{c}} & Maximum feature maps & 100  \\
\textit{r} & Dropout rate & 0.5 \\ 
\textit{e} & Maximum epochs & 50  \\ 
\textit{mb} & Mini-batch size & 50  \\
\textit{$\lambda$} & Learning rate & 0.2  \\
\textit{$\alpha_1$} & Threshold to filter word TF-IDF & 7  \\ 
\textit{$\alpha_2$} & Threshold to filter Pair TF-IDF & 5  \\ 
\textit{$\alpha_3$} & Threshold to filter Triplet TF-IDF & 2 \\
\textit{m} & Threshold for sub-graph alignment & 3  \\
\hline
\end{tabular}

\label{param}
\end{table}
\begin{table*}[]
\centering
\caption{Evaluation results of answer triggering on the development and test set of
\textsc{WikiQA} dataset: Question-level precision, recall and F-scores. MAP and MRR are used to evaluate the performance of answer selection.
\textbf{FMap} denotes the size of feature map. Precision, Recall and F-score are given in percentages(\%).}
\resizebox{\textwidth}{!}{%
\begin{tabular}{l|l|c|c|c|c|c|c|c|l|l}
\hline
&
\multicolumn{5}{|c}{\textbf{Test set}} &
\multicolumn{5}{|c}{\textbf{Development set}}
\\  \hline
\hline
\textbf{Model} & \textbf{MAP} & \textbf{MRR
}& \textbf{Precision} & \textbf{Recall} & \textbf{F-score} & \textbf{MAP} & \textbf{MRR
}& \textbf{Precision} & \textbf{Recall} & \textbf{F-score}\\ \hline
\multicolumn{11}{c}{\textbf{Baselines}}\\ \hline \hline
BM-25 & $0.4712$  & $0.4889$ & $21.04$ & $24.43$  & $22.60$ &

$0.4569$  & $0.4701$ & $20.32$ & $26.19$  & $22.88$  \\ \hline
N-Gram Coverage & $0.5102$  & $0.5349$ & $24.47$ & $28.59$  & $25.24$  &

$0.5289$  & $0.4909$ & $25.73$ & $29.11$  & $27.31$\\ \hline

Semantic Vec (W2V) & $0.4323$ & $0.4411$ & $13.37$ & $34.16$ & $19.21$

& $0.4429$  & $0.4395$ & $14.55$ & $35.87$  & $20.70$
\\ \hline
Semantic Vec (Glove) & $0.4928$ & $0.5277$ & $16.12$ & $40.74$ & $23.10$ 

& $0.4987$  & $0.4901$ & $17.56$ & $39.19$  & $24.25$

\\ \hline

\multicolumn{11}{c}{\textbf{Our Models}} \\ \hline \hline
Joint-CNN (50-FMap)& $0.6218$ & $0.6322$ & $28.90$ & $31.28$ & $30.04$ 

& $0.6123$  & $0.6520$ & $33.01$ & $26.98$  & $29.69$
\\ \hline
Joint-CNN (150-FMap)& $0.6369$ & $0.6535$ & $25.07$ & $39.51$ & $30.67$ 

& $0.6152$  & $0.6245$ & $25.29$ & $34.13$  & $29.05$
\\ \hline
Joint-CNN (100-FMap)& $0.6372$ & $0.6567$ & $23.21$ & $\mathbf{48.15}$ & $31.33$ 

& $0.6220$  & $0.6250$ & $25.33$ & $46.03$  & $32.68$

\\ \hline
+ Graph Edit Distance & $0.6540$ & $0.6708$ & $33.18$ & $30.04$ & $31.53$ 

& $0.6487$  & $0.6522$ & $32.53$ & $36.28$  & $34.30$
\\ \hline
+ Graph Similarity & $0.6648$ & $0.6828$ & $25.00$ & $43.21$ & $31.67$ 

& $0.6810$  & $0.6744$ & $34.85$ & $36.51$  & $35.66$
\\ \hline
+ Graph Coverage & $\mathbf{0.6793}$ & $\mathbf{0.6908}$ & $\mathbf{35.69}$ & $39.51$ & $\mathbf{37.50}$ 

& $0.6917$  & $0.6940$ & $39.83$ & $37.30$  & $38.52$
\\ \hline
\multicolumn{11}{c}{\textbf{State-of-the art (Answer Triggering)}}                          \\ \hline \hline
\begin{tabular}[c]{@{}l@{}}CNN-cnt+All\\ \cite{yang-yih-meek:2015:EMNLP} \end{tabular} & $-$ & $-$ & $28.34$ & $35.80$ & $31.64$ 

& $-$  & $-$ & $-$ & $-$  & $-$
\\ \hline
\begin{tabular}[c]{@{}l@{}}CNN$_{3}$: max + emb+\\ \cite{selQAPaper} \end{tabular} & $-$ & $-$ & $29.43$ & $48.56$ & $36.65$ 

& $-$  & $-$ & $-$ & $-$  & $-$
\\ \hline
\end{tabular}
}

\label{result}
\end{table*}
\begin{table*}[]
\centering
\caption{Comparative analysis of the result from Joint-CNN and Hybrid model on pair of questions and answers. The correct answer are in \textbf{bold}. Here \textbf{TG}: trigger and \textbf{NTG}:non-trigger model predictions, marked as correct `\ding{51}' or incorrect `\ding{53}'.}
\resizebox{\textwidth}{!}{%
\begin{tabular}{l|l|c|c}
\hline
\multicolumn{1}{c|}{\textbf{Question}} & \multicolumn{1}{c|}{\textbf{Answer}} & \textbf{Joint-CNN} & \textbf{Hybrid} \\ \hline
\multicolumn{1}{c|}{\multirow{2}{*}{What is adoration catholic church ?}} & \begin{tabular}[c]{@{}l@{}}Adoration is a sign of devotion to and \\ worship of Jesus Christ , who is believed \\ by Catholics to be present Body, Blood,\\  Soul, and Divinity.\end{tabular} & \begin{tabular}[c]{@{}c@{}}TG\\ (\ding{53})\end{tabular} &\begin{tabular}[c]{@{}c@{}}NTG\\ (\ding{51})\end{tabular} \\ \cline{2-4} 
\multicolumn{1}{c|}{} & \textbf{\begin{tabular}[c]{@{}l@{}}Eucharistic adoration is a practice in the\\  Roman Catholic Church , and in a few \\ Anglican and Lutheran churches, in which \\ the Blessed Sacrament is exposed and adored \\ by the faithful.\end{tabular}} & \begin{tabular}[c]{@{}c@{}}NTG\\ (\ding{53})\end{tabular} & \begin{tabular}[c]{@{}c@{}}TG\\ (\ding{51})\end{tabular} \\ \hline
\multirow{2}{*}{where is La Palma africa ?} & \begin{tabular}[c]{@{}l@{}}La Palma has an area of 706 km making it\\  the fifth largest of the seven main Canary \\ Islands.\end{tabular} & \begin{tabular}[c]{@{}c@{}}TG\\ (\ding{53})\end{tabular} &\begin{tabular}[c]{@{}c@{}}NTG\\ (\ding{51})\end{tabular} \\ \cline{2-4} 
 & \textbf{\begin{tabular}[c]{@{}l@{}}La Palma is the most north-westerly of\\  the Canary Islands.\end{tabular}} & \begin{tabular}[c]{@{}c@{}}NTG\\ (\ding{53})\end{tabular} & \begin{tabular}[c]{@{}c@{}}TG\\ (\ding{51})\end{tabular} \\ \hline
\multirow{2}{*}{What are land parcels} & \textbf{land lot, a piece of land;} & \begin{tabular}[c]{@{}c@{}}NTG\\ (\ding{53})\end{tabular} & \begin{tabular}[c]{@{}c@{}}TG\\ (\ding{51})\end{tabular} \\ \cline{2-4} 
 & fluid parcel, a concept in fluid dynamics & TG & NTG \\ \hline
What bacteria grow on macconkey agar & \textbf{\begin{tabular}[c]{@{}l@{}}MacConkey agar is a culture medium\\  designed to grow Gram-negative bacteria\\  and differentiate them for lactose fermentation .\end{tabular}} & \begin{tabular}[c]{@{}c@{}}TG\\ (\ding{51})\end{tabular} &\begin{tabular}[c]{@{}c@{}}NTG\\ (\ding{53})\end{tabular} \\ \hline
How much is centavos in Mexico & \textbf{The peso is subdivided into 100 centavos.} & \begin{tabular}[c]{@{}c@{}}NTG\\ (\ding{53})\end{tabular} &\begin{tabular}[c]{@{}c@{}}NTG\\ (\ding{53})\end{tabular} \\ \hline
\end{tabular}%
}

\label{analysis}
\end{table*}
\begin{itemize}
\item \textbf{Q:} \textit{What is adoration catholic church ?
}\\
$\mathbf{A_{1}:}$ \textit{Adoration is a sign of devotion to and worship of Jesus Christ , who is believed by Catholics to be present Body, Blood, Soul, and Divinity.
}\\
$\mathbf{A_{2}:}$ \textit{Eucharistic adoration is a practice in the Roman Catholic Church , and in a few Anglican and Lutheran churches, in which the Blessed Sacrament is exposed and adored by the faithful.
}\\
Here, the Joint-CNN model selects the answer as $\mathbf{A_{1}}$, but the hybrid model selects $\mathbf{A_{2}}$, which is correct. The reason could be that vocab and graph coverage \footnote{\textit{church} and	\textit{adoration} are common node between question and answer which increases the graph coverage} between \textbf{Q} and $\mathbf{A_{2}}$ are higher than \textbf{Q} and $\mathbf{A_{1}}$.

\item \textbf{Q:} \textit{where is La Palma africa ?
}\\
$\mathbf{A_{1}:}$ \textit{La Palma has an area of 706 km2 making it the fifth largest of the seven main Canary Islands.
}\\
$\mathbf{A_{2}:}$ \textit{La Palma is the most north-westerly of the Canary Islands.
}\\
Both $\mathbf{A_{1}}$ and $\mathbf{A_{2}}$ are the correct answers for the question $\mathbf{Q}$, but $\mathbf{A_{2}}$ has more precise information compared to $\mathbf{A_{1}}$. Joint-CNN model selects $\mathbf{A_{1}}$ as correct answer, whereas the hybrid model selects $\mathbf{A_{2}}$, because of the higher graph coverage score.

\end{itemize}
\section{Conclusion}
In this paper, we have proposed a hybrid model for answer triggering using deep learning and graph based features. Modeling QA pair is a more complex task than classifying a single sentence. It is observed that CNN does not capture the important features spanning between the text such as quantification of similarity/dissimilarity, geometric similarity. To overcome this limitation, we investigate the incorporation of richer linguistic features (dependency graph) in CNN. Experiments on the \textsc{WikiQA} benchmark dataset show that integrating graph-based alignment features with CNNs improves the performance significantly. Future work includes to build a end to end neural network which can embedded graph based feature with sentence encoder (CNN, LSTM etc.)
\bibliographystyle{splncs03}
\bibliography{cicling} 

\begin{thebibliography}{10}
\providecommand{\url}[1]{\texttt{#1}}
\providecommand{\urlprefix}{URL }

\bibitem{albalate2013semi}
Albalate, A., Minker, W.: Semi-Supervised and Unervised Machine Learning: Novel
  Strategies. John Wiley \& Sons (2013)

\bibitem{kalchbrenner2014convolutional}
Blunsom, P., Grefenstette, E., Kalchbrenner, N.: A convolutional neural network
  for modelling sentences. In: Proceedings of the 52nd Annual Meeting of the
  Association for Computational Linguistics. Proceedings of the 52nd Annual
  Meeting of the Association for Computational Linguistics (2014)

\bibitem{collobert2011natural}
Collobert, R., Weston, J., Bottou, L., Karlen, M., Kavukcuoglu, K., Kuksa, P.:
  Natural language processing (almost) from scratch. Journal of Machine
  Learning Research  12(Aug),  2493--2537 (2011)

\bibitem{dijkstra1959note}
Dijkstra, E.W.: A note on two problems in connexion with graphs. Numerische
  mathematik  1(1),  269--271 (1959)

\bibitem{dong2015question}
Dong, L., Wei, F., Zhou, M., Xu, K.: Question answering over freebase with
  multi-column convolutional neural networks. In: ACL (1). pp. 260--269 (2015)

\bibitem{DBLP:journals/corr/FengXGWZ15}
Feng, M., Xiang, B., Glass, M.R., Wang, L., Zhou, B.: Applying deep learning to
  answer selection: A study and an open task. In: Automatic Speech Recognition
  and Understanding (ASRU), 2015 IEEE Workshop on. pp. 813--820. IEEE (2015)

\bibitem{graff2003english}
Graff, D., Cieri, C.: English gigaword, ldc catalog no. LDC2003T05. Linguistic
  Data Consortium, University of Pennsylvania  (2003)

\bibitem{GUPTA18.486}
Gupta, D., Ekbal, A., Bhattacharyya, P.: {A Deep Neural Network based Approach
  for Entity Extraction in Code-Mixed Indian Social Media Text}. In: chair),
  N.C.C., Choukri, K., Cieri, C., Declerck, T., Goggi, S., Hasida, K., Isahara,
  H., Maegaard, B., Mariani, J., Mazo, H., Moreno, A., Odijk, J., Piperidis,
  S., Tokunaga, T. (eds.) Proceedings of the Eleventh International Conference
  on Language Resources and Evaluation (LREC 2018). European Language Resources
  Association (ELRA), Miyazaki, Japan (May 7-12, 2018 2018)

\bibitem{GUPTA18.826}
Gupta, D., Kumari, S., Ekbal, A., Bhattacharyya, P.: {MMQA: A Multi-domain
  Multi-lingual Question-Answering Framework for English and Hindi}. In:
  chair), N.C.C., Choukri, K., Cieri, C., Declerck, T., Goggi, S., Hasida, K.,
  Isahara, H., Maegaard, B., Mariani, J., Mazo, H., Moreno, A., Odijk, J.,
  Piperidis, S., Tokunaga, T. (eds.) Proceedings of the Eleventh International
  Conference on Language Resources and Evaluation (LREC 2018). European
  Language Resources Association (ELRA), Miyazaki, Japan (May 7-12, 2018 2018)

\bibitem{gupta-EtAl:2016:W16-63}
Gupta, D., Lamba, A., Ekbal, A., Bhattacharyya, P.: Opinion mining in a
  code-mixed environment: A case study with government portals. In: Proceedings
  of the 13th International Conference on Natural Language Processing. pp.
  249--258. NLP Association of India, Varanasi, India (December 2016),
  \url{http://www.aclweb.org/anthology/W/W16/W16-6331}

\bibitem{hecht1989theory}
Hecht-Nielsen, R.: Theory of the backpropagation neural network. In: Neural
  Networks, 1989. IJCNN., International Joint Conference on. pp. 593--605. IEEE
  (1989)

\bibitem{heilman2010tree}
Heilman, M., Smith, N.A.: Tree edit models for recognizing textual entailments,
  paraphrases, and answers to questions. In: Human Language Technologies: The
  2010 Annual Conference of the North American Chapter of the Association for
  Computational Linguistics. pp. 1011--1019. Association for Computational
  Linguistics (2010)

\bibitem{iyyer-EtAl:2014:EMNLP2014}
Iyyer, M., Boyd-Graber, J., Claudino, L., Socher, R., Daum\'{e}~III, H.: A
  neural network for factoid question answering over paragraphs. In:
  Proceedings of the 2014 Conference on Empirical Methods in Natural Language
  Processing (EMNLP). pp. 633--644. Association for Computational Linguistics,
  Doha, Qatar (October 2014), \url{http://www.aclweb.org/anthology/D14-1070}

\bibitem{selQAPaper}
Jurczyk, T., Zhai, M., Choi, J.D.: Selqa: A new benchmark for selection-based
  question answering. In: 2016 IEEE 28th International Conference on Tools with
  Artificial Intelligence (ICTAI). pp. 820--827 (Nov 2016)

\bibitem{kalchbrenner-grefenstette-blunsom:2014:P14-1}
Kalchbrenner, N., Grefenstette, E., Blunsom, P.: A convolutional neural network
  for modelling sentences. In: Proceedings of the 52nd Annual Meeting of the
  Association for Computational Linguistics (Volume 1: Long Papers). pp.
  655--665. Association for Computational Linguistics, Baltimore, Maryland
  (June 2014), \url{http://www.aclweb.org/anthology/P14-1062}

\bibitem{kim2014convolutional}
Kim, Y.: Convolutional neural networks for sentence classification. In:
  Proceedings of the 2014 Conference on Empirical Methods in Natural Language
  Processing (EMNLP). pp. 1746--1751. Association for Computational
  Linguistics, Doha, Qatar (October 2014),
  \url{http://www.aclweb.org/anthology/D14-1181}

\bibitem{kohail2015unsupervised}
Kohail, S.: Unsupervised topic-specific domain dependency graphs for aspect
  identification in sentiment analysis. In: Proceedings of the Student Research
  Workshop associated with RANLP. pp. 16--23 (2015)

\bibitem{kohail2017cicling}
Kohail, S., Biemann, C.: Matching, re-ranking and scoring: Learning textual
  similarity by incorporating dependency graph alignment and coverage features.
  In: 18th International Conference on Computational Linguistics and
  Intelligent Text Processing (2017)

\bibitem{kumar2016recurrent}
Kumar, A., Ekbal, A., Saha, S., Bhattacharyya, P., et~al.: A recurrent neural
  network architecture for de-identifying clinical records. In: Proceedings of
  the 13th International Conference on Natural Language Processing. pp.
  188--197 (2016)

\bibitem{miao2016neural}
Miao, Y., Yu, L., Blunsom, P.: Neural variational inference for text
  processing. In: International Conference on Machine Learning. pp. 1727--1736
  (2016)

\bibitem{munkres1957algorithms}
Munkres, J.: Algorithms for the assignment and transportation problems. Journal
  of the society for industrial and applied mathematics  5(1),  32--38 (1957)

\bibitem{ren2015exploring}
Ren, M., Kiros, R., Zemel, R.: Exploring models and data for image question
  answering. In: Advances in neural information processing systems. pp.
  2953--2961 (2015)

\bibitem{robertson1995okapi}
Robertson, S.E., Walker, S., Jones, S., Hancock-Beaulieu, M.M., Gatford, M.,
  et~al.: Okapi at trec-3. NIST SPECIAL PUBLICATION SP  109,  109 (1995)

\bibitem{salton1988term}
Salton, G., Buckley, C.: Term-weighting approaches in automatic text retrieval.
  Information processing \& management  24(5),  513--523 (1988)

\bibitem{dos2015boosting}
dos Santos, C., Guimaraes, V., Niter{\'o}i, R., de~Janeiro, R.: Boosting named
  entity recognition with neural character embeddings. In: Proceedings of NEWS
  2015 The Fifth Named Entities Workshop. p.~25 (2015)

\bibitem{severyn2013automatic}
Severyn, A., Moschitti, A.: Automatic feature engineering for answer selection
  and extraction. In: EMNLP. pp. 458--467 (2013)

\bibitem{2016deep}
Shweta, A.E., Saha, S., Bhattacharyya, P.: Deep learning architecture for
  patient data de-identification in clinical records. In: Proceeding of
  Clinical Natural Language Processing Workshop (ClinicalNLP) at the 26th
  International Conference on Computational Linguistics (COLING 2016), Japan
  (accepted) (2016)

\bibitem{socher2013parsing}
Socher, R., Bauer, J., Manning, C.D., Ng, A.Y.: Parsing with compositional
  vector grammars. In: In Proceedings of the ACL conference. Citeseer (2013)

\bibitem{socher2013recursive}
Socher, R., Perelygin, A., Wu, J.Y., Chuang, J., Manning, C.D., Ng, A.Y.,
  Potts, C.: Recursive deep models for semantic compositionality over a
  sentiment treebank. In: Proceedings of the conference on empirical methods in
  natural language processing (EMNLP). vol. 1631, p. 1642. Citeseer (2013)

\bibitem{turian2010word}
Turian, J., Ratinov, L., Bengio, Y.: Word representations: a simple and general
  method for semi-supervised learning. In: Proceedings of the 48th annual
  meeting of the association for computational linguistics. pp. 384--394.
  Association for Computational Linguistics (2010)

\bibitem{DBLP:journals/corr/WanLGXPC15}
Wan, S., Lan, Y., Guo, J., Xu, J., Pang, L., Cheng, X.: A deep architecture for
  semantic matching with multiple positional sentence representations. CoRR
  abs/1511.08277 (2015)

\bibitem{xiong2016dynamic}
Xiong, C., Merity, S., Socher, R.: Dynamic memory networks for visual and
  textual question answering. In: International Conference on Machine Learning.
  pp. 2397--2406 (2016)

\bibitem{lrec2018}
Yadav, S., Ekbal, A., Saha, S., Bhattacharyya, P.: {Medical Sentiment Analysis
  using Social Media: Towards building a Patient Assisted System}. In: chair),
  N.C.C., Choukri, K., Cieri, C., Declerck, T., Goggi, S., Hasida, K., Isahara,
  H., Maegaard, B., Mariani, J., Mazo, H., Moreno, A., Odijk, J., Piperidis,
  S., Tokunaga, T. (eds.) Proceedings of the Eleventh International Conference
  on Language Resources and Evaluation (LREC 2018). European Language Resources
  Association (ELRA), Miyazaki, Japan (May 7-12, 2018 2018)

\bibitem{yang-yih-meek:2015:EMNLP}
Yang, Y., Yih, W.t., Meek, C.: Wikiqa: A challenge dataset for open-domain
  question answering. In: Proceedings of the 2015 Conference on Empirical
  Methods in Natural Language Processing. pp. 2013--2018. Association for
  Computational Linguistics, Lisbon, Portugal (September 2015)

\bibitem{yao2013answer}
Yao, X., Van~Durme, B., Callison-Burch, C., Clark, P.: Answer extraction as
  sequence tagging with tree edit distance. In: HLT-NAACL. pp. 858--867 (2013)

\bibitem{yih-he-meek:2014:P14-2}
Yih, W.t., He, X., Meek, C.: Semantic parsing for single-relation question
  answering. In: Proceedings of the 52nd Annual Meeting of the Association for
  Computational Linguistics (Volume 2: Short Papers). pp. 643--648. Association
  for Computational Linguistics, Baltimore, Maryland (June 2014),
  \url{http://www.aclweb.org/anthology/P14-2105}

\bibitem{yin2016attention}
Yin, W., Ebert, S., Sch{\"u}tze, H.: Attention-based convolutional neural
  network for machine comprehension. arXiv preprint arXiv:1602.04341  (2016)

\bibitem{TACL831}
Yin, W., Schütze, H., Xiang, B., Zhou, B.: Abcnn: Attention-based
  convolutional neural network for modeling sentence pairs. Transactions of the
  Association for Computational Linguistics  4,  259--272 (2016),
  \url{https://transacl.org/ojs/index.php/tacl/article/view/831}

\bibitem{yu2014deep}
Yu, L., Hermann, K.M., Blunsom, P., Pulman, S.: Deep learning for answer
  sentence selection. arXiv preprint arXiv:1412.1632  (2014)

\bibitem{zeiler2012adadelta}
Zeiler, M.D.: {ADADELTA:} an adaptive learning rate method. CoRR  abs/1212.5701
  (2012), \url{http://arxiv.org/abs/1212.5701}

\end{thebibliography}
\end{document}